\documentclass[runningheads]{llncs}
\usepackage[T1]{fontenc}
\usepackage{graphicx}
\usepackage{color}
\usepackage{hyperref}
\hypersetup{hidelinks, colorlinks,breaklinks=true,urlcolor=black,citecolor=black,linkcolor=black}

\usepackage{booktabs}
\usepackage[table]{xcolor}
\usepackage{amssymb}
\usepackage{subcaption}
\usepackage{amsmath}
\usepackage{amssymb}
\usepackage{multicol}
\usepackage{multirow}
\usepackage{xspace}

\DeclareMathOperator*{\argmax}{arg\,max}

\begin{document}
\title{Conditioned Prompt-Optimization for Continual
Deepfake Detection} %
\author{Francesco Laiti\inst{1} \and
Benedetta Liberatori\inst{1} \and
Thomas De Min\inst{1} \and
Elisa Ricci\inst{1,2}
}
\authorrunning{F. Laiti et al.}
\institute{University of Trento \and 
Fondazione Bruno Kessler
}
\maketitle              %

\newcommand{\methodname}{Prompt2Guard\xspace}
\def\ie{\textit{i.e.}}
\def\eg{\textit{e.g.}}
\newcommand{\thomas}[1]{\textcolor{blue}{[Thomas: #1]}}
\newcommand{\benni}[1]{\textcolor{magenta}{[B: #1]}}
\newcommand{\franz}[1]{\textcolor{orange}{[F: #1]}}

\newcommand{\dataset}[0]{\mathcal{D}}

\definecolor{TableBlue}{RGB}{218,232,252}
\definecolor{TableGreen}{RGB}{213,232,212}
\definecolor{TablePink}{RGB}{250,218,221}

\definecolor{TextRed}{RGB}{204,93,89}
\definecolor{TextGreen}{RGB}{135,186,106}
\definecolor{TextBlue}{RGB}{0,113,227}

\newcommand{\inlineColorbox}[2]{\begingroup\setlength{\fboxsep}{1pt}\colorbox{#1}{\hspace*{2pt}\vphantom{Ay}#2\hspace*{2pt}}\endgroup}

\begin{abstract}

The rapid advancement of generative models has significantly enhanced the realism and customization of digital content creation. 
The increasing power of these tools, coupled with their ease of access, fuels the creation of photorealistic fake content, termed deepfakes, that raises substantial concerns about their potential misuse. 
In response, there has been notable progress in developing detection mechanisms to identify content produced by these advanced systems. 
However, existing methods often struggle to adapt to the continuously evolving landscape of deepfake generation.
This paper introduces \methodname, a novel solution for exemplar-free continual deepfake detection of images, that leverages Vision-Language Models (VLMs) and domain-specific multi-modal prompts. 
Compared to previous VLM-based approaches that are either bounded by prompt selection accuracy or necessitate multiple forward passes, we leverage a prediction ensembling technique with read-only prompts.
Read-only prompts do not interact with VLMs internal representation, mitigating the need for multiple forward passes. 
Thus, we enhance efficiency and accuracy in detecting generated content. 
Additionally, our method exploits a text-prompt conditioning tailored to deepfake detection, which we demonstrate is beneficial in our setting.
We evaluate \methodname~on CDDB-Hard, a continual deepfake detection benchmark composed of five deepfake detection datasets spanning multiple domains and generators, achieving a new state-of-the-art. %
Additionally, our results underscore the effectiveness of our approach in addressing the challenges posed by continual deepfake detection, paving the way for more robust and adaptable solutions in deepfake detection. Source code is available at \url{https://github.com/laitifranz/Prompt2Guard}.

\end{abstract}

\section{Introduction}\label{sec:intro}

The rapid evolution of generative artificial intelligence has revolutionized the digital content creation, enabling unprecedented levels of realism, customization, and accuracy~\cite{ramesh2022hierarchical,rombach2022high,blattmann2023align,ge2023preserve}. The ease of access to these technologies has been crucial in driving their advancement, making powerful tools available to a broader audience beyond researchers, thereby removing barriers for non-expert users. 
This progress has led to photorealistic fake images and videos, \ie{} \textit{deepfakes}, raising significant concerns regarding their potential for malicious use. With human discernment facing significant challenges in distinguishing between real and generated fake images~\cite{NEURIPS2023_505df5ea}, urgent attention is needed to develop effective detection mechanisms capable of accurately identifying content produced by these advanced systems. 

\begin{figure}[t]
    \centering
    \begin{subfigure}{0.49\linewidth}
        \includegraphics[width=\linewidth]{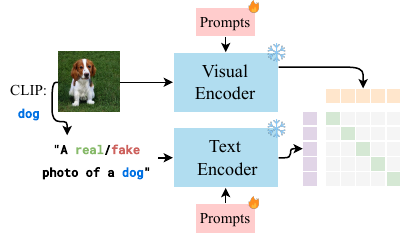}
        \caption{Training}
        \label{fig:teaser_train}
    \end{subfigure}%
    \begin{subfigure}{0.49\linewidth}
        \includegraphics[width=\linewidth]{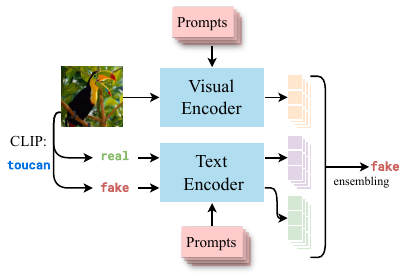}
        \caption{Inference}
        \label{fig:teaser_test}
    \end{subfigure}
    \caption{\textbf{Overview of the proposed method.} \methodname addresses the task of domain incremental deepfake detection. The training (a) is performed on a sequence of datasets, coming from different domains. At inference time (b) the model classifies the input image into real or fake, without domain knowledge.}
    \label{fig:enter-label}
\end{figure}

Significant progress has been achieved on the deepfake detection as well, with state-of-the-art detectors capable of identifying images generated using GANs and diffusion models~\cite{bird2024cifake,10.1145/3576915.3616588}. However, these methods primarily function within a stationary scenario, wherein a large amount of relatively homogeneous deepfake content is presented at training time. This ideal scenario is often not reflective of the real-world landscape. In practice, likely heterogeneous deepfakes are continuously produced using novel and unseen architectures, presenting a constantly evolving landscape for detection methods to navigate. To effectively tackle this challenge, modern deepfake detectors must be able to adapt to the latest generators without succumbing to catastrophic forgetting. Maintaining the ability to detect content from diverse generators is crucial, as older generators continue to pose a significant threat. 

While some continual deepfake detection benchmarks have been introduced lately~\cite{li2023continual,genimage23}, the field still lacks comprehensive exploration, with few methods tackling deepfake detection in the incremental setting. Capitalizing on the generalization abilities of Vision-Language Models (VLMs), recent methods have shown promise in leveraging the encoded knowledge of these models for deepfake detection~\cite{wang2022sprompt,Nicolas_2024_WACV}. 

These methods adapt VLMs to Domain Incremental Learning (DIL) by learning specific prompts for each task (generator) at training time, thereby maintaining independence in the training process. At test time, they either require inferring the generator to select the appropriate prompts~\cite{wang2022sprompt} or, when undecided, performing multiple forward passes and aggregate information from different parameters~\cite{Nicolas_2024_WACV}.
As a result, they are either constrained by task selection accuracy or necessitate expensive multiple forwards to output a single prediction. 
Furthermore, while VLMs demonstrate potential in assessing the authenticity of visual content, their application in deepfake detection often oversimplifies the problem, treating the task as standard binary classification. 

Inspired by these observations, we propose~\methodname, a novel solution for examplar-free continual deepfake detection that leverages VLMs and domain-specific multi-modal prompts, as illustrated in Fig.~\ref{fig:enter-label}. Compared to previous VLM-based methods, our solution is specifically tailored for deepfakes and solves the task selection problem with a prediction ensembling that does not require multiple forward passes. We evaluate the proposed approach on the challenging CDDB benchmark~\cite{li2023continual}, consisting of a sequence of datasets coming from different image generators, achieving state-of-the-art results. 

Our contributions can be summarized as follows:
\begin{enumerate}
    \item We present~\methodname, a novel VLM-based exemplar-free DIL strategy that leverages multi-modal prompts. These prompts are read-only and do not alter the VLM internal representation. As a result, we can ensemble prediction scores from different tasks without requiring multiple forward passes, enhancing accuracy and efficiency. 
    \item Additionally, we propose a text-prompt conditioning procedure specifically tailored to deepfake detection and show its effectiveness. 
    \item We empirically show the capabilities of the proposed method, achieving state-of-the-art results in task-wise average accuracy without incurring catastrophic forgetting on the CDDB benchmark~\cite{li2023continual}. 
\end{enumerate}

\section{Related work}\label{sec:related}

\noindent\textbf{Deepfake Detection.}
The field of media forensics has a long history of utilizing traditional tools to analyze synthetic images, including techniques such as identifying resampling artifacts~\cite{1381775}, JPEG quantization~\cite{8267641}, image splicing~\cite{huh2018fighting}, and Photoshop warping~\cite{wang2019detecting}.
With the democratization of synthetic image creation through deep generative methods, recent studies have focused on employing deep discriminative methods to detect such manipulated content, particularly for GAN-based approaches~\cite{chai2020makes}. Rössler et al.~\cite{roessler2019} train an Xception~\cite{chollet2016xception} for detecting deepfake images of faces. Chai et al.~\cite{chai2020makes} employ limited receptive fields to identify the most indicative patches, demonstrating that they contain sufficient cues for detecting images as real or fake. Wang et al.~\cite{wang2020cnn} show that CNN-generated images share common flaws and a ResNet-50~\cite{He2015DeepRL} with suitable data augmentations can generalize across generators.

\noindent\textbf{Vision-language models.} Vision-language models (VLMs), pioneered by CLIP ~\cite{radford2021learning}, are pre-trained on a vast amount of web-crawled image-text pairs to learn joint visual-text embedding spaces. 
These models have demonstrated outstanding performance in various downstream tasks, especially in zero-shot image classification~\cite{jia2021scaling,yu2022coca}. While VLMs exhibit robust generalization capabilities, adapting them to specific tasks is challenging. Recent studies in VLMs involve prompt learning to adapt pre-trained models to downstream tasks using affordable-sized datasets. CoOp~\cite{zhou2022learning} uses continuous vector prompts, which are concatenated and processed with text tokens. CoCoOp~\cite{zhou2022conditional} further extends CoOp by leveraging a lightweight neural network to generate prompts conditioned on the input image.  
These works keep the pre-trained weights frozen, yet the learnable prompts still affect the model’s hidden representation through the attention mechanism. To prevent this internal representation shift, RPO~\cite{lee2023read} proposes to use a masked attention mechanism, which limits prompts to only read information from the attention-based interactions of the pre-trained model.

\noindent \textbf{Continual Learning.} 
To tackle catastrophic forgetting, early continual learning approaches proposed regularization terms to constrain network parameters from forgetting old knowledge when updated with new information~\cite{li2017learning,kirkpatrick2017overcoming,aljundi2018memory,chaudhry2018riemannian,zenke2017continual,castro2018end,fini2022self}.
Despite reducing forgetting during sequential updates, these methods fail to retain satisfactory performances after multiple updates.
By allowing the storage of part of the old data in memory buffers, rehearsal-based approaches~\cite{rebuffi2017icarl,buzzega2020dark,wu2019large,prabhu2020gdumb,cha2021co2l} have shown superior performance over memory-free approaches.
However, storing data in a replay buffer for future rehearsal poses privacy-related concerns as data may be leaked.
To preserve privacy while maintaining performances similar to replay-based methods, parameter-isolation approaches exploit pre-trained models and tune just a small fraction of parameters for each update~\cite{wang2022learning,wang2022dualprompt,smith2023coda,de2023effectiveness,zhou2023revisiting,mcdonnell2024ranpac}, which are selected at inference time through a query-key selection mechanism.
In particular, S-Prompts~\cite{wang2022sprompt} tackles the domain-incremental learning problem (DIL) by tuning CLIP pre-training~\cite{radford2021learning} on domain-independent sets of vision and language prompts.
Since the domain shift across incremental steps is high in DIL, the query-key matching likely targets the best set of tuned prompts.
MoP-CLIP~\cite{Nicolas_2024_WACV}, instead, proposes a mixture of prompt-tuned CLIP models to address the poor out-of-distribution performance of S-Prompts and DIL methods.

In this work, we propose to tune CLIP vision and language encoders with multi-modal read-only prompts~\cite{lee2023read}, which
allow us to compute multiple and parallel representations for each image, thus, computing the final prediction as a weighted sum of domain-specific predictions.

\section{Preliminaries}
\label{sec:preliminaries}

\subsection{Problem formulation}
\label{sub:problem}
In deepfake detection, the objective is to train a model capable of distinguishing real images from synthetically generated ones.
Formally the model $f_\theta: \mathcal{X} \rightarrow \mathcal{Y}$, parameterized by $\theta$, maps images from the input space $\mathcal{X}$ to the binary semantic space $\mathcal{Y} = \{0, 1\}$, where generated samples should be predicted as 1.
In this work, we tackle the problem of deepfake detection in an incremental learning scenario, a particular instance of domain incremental learning, where $f_\theta$ must be trained sequentially over non-stationary datasets.
Let $\dataset = \{\dataset^1, ..., \dataset^T\}$ be the sequence of datasets, the $k$-th dataset $\dataset^k = \{(x_i, y_i)\}_{i=1}^{N_k}$ is composed of real and generated images with their corresponding semantic annotation.
At each step, synthetic images are generated with a different generator $\mathcal{G}^k$, thus, the input data distribution shifts from task to task.
Given two distinct tasks $k$ and $m$, with $k\neq m$, the distributions of the two tasks are different, \ie{} $p(\mathcal{X}^k) \neq p(\mathcal{X}^m)$.   
Given a new domain, DIL aims to improve the model’s performance on the latest distribution, while avoiding the loss of knowledge for past domains. 
At inference time, the model must classify the input image without knowing the domain.
In the following, we will interchangeably use the terms \textit{domain} and \textit{task}.

\subsection{Prompt tuning}
\label{sub:fine-tuning}
We follow previous works in the field~\cite{wang2022sprompt,Nicolas_2024_WACV} and fine-tune CLIP~\cite{radford2021learning} on the sequential datasets.
Previous works in DIL~\cite{wang2022sprompt,Nicolas_2024_WACV} exploit prompt tuning for adapting CLIP to the incremental detection of deepfakes.
As prompts are specific for particular types of generated data, the training procedure of these approaches is independent for each generator.
This reduces the risk of forgetting, as prompts, once trained, are kept frozen throughout the entire lifetime of the model, creating distinct subspaces for each domain's knowledge rather than relying on a shared feature space for all tasks, thereby reducing the interference between old and new domains.
However, training task-specific prompts forces previous methods to guess prompts to use at inference time, and this operation must be performed for each image.
Images are then selected using the guessed prompts.
This limits previous methods since wrong prompt selection results in lower model accuracy.
MoP-CLIP~\cite{Nicolas_2024_WACV} solves this issue by forwarding the target image multiple times with different trained prompts when the query-key selection mechanism has low confidence.
However, this means MoP-CLIP has to forward the target image as many times as the number of tasks encountered by the model, which does not scale well in practice.

\section{Prompt2Guard}\label{sec:method}

\begin{figure}
    \centering
    \includegraphics[width=0.8\linewidth]{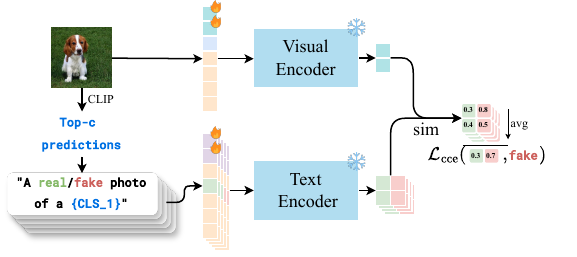}
    \caption{\textbf{Illustration of the training.} The prepended prompts are the only learnable parameters (\raisebox{-1.3mm}{\includegraphics[height=4.3mm]{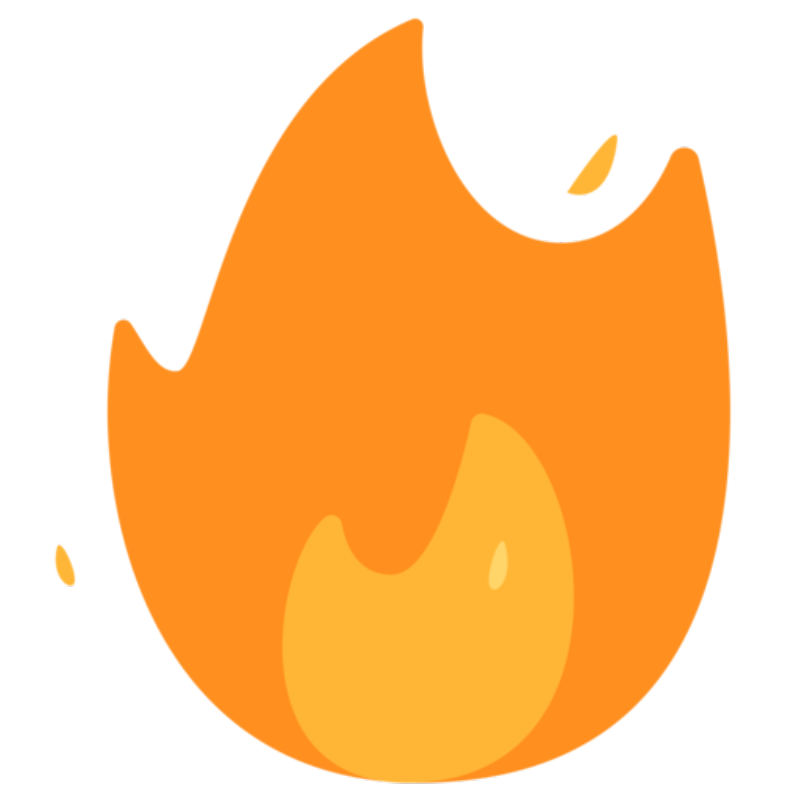}}), while the encoders are kept frozen (\raisebox{-1.3mm}{\includegraphics[height=4.3mm]{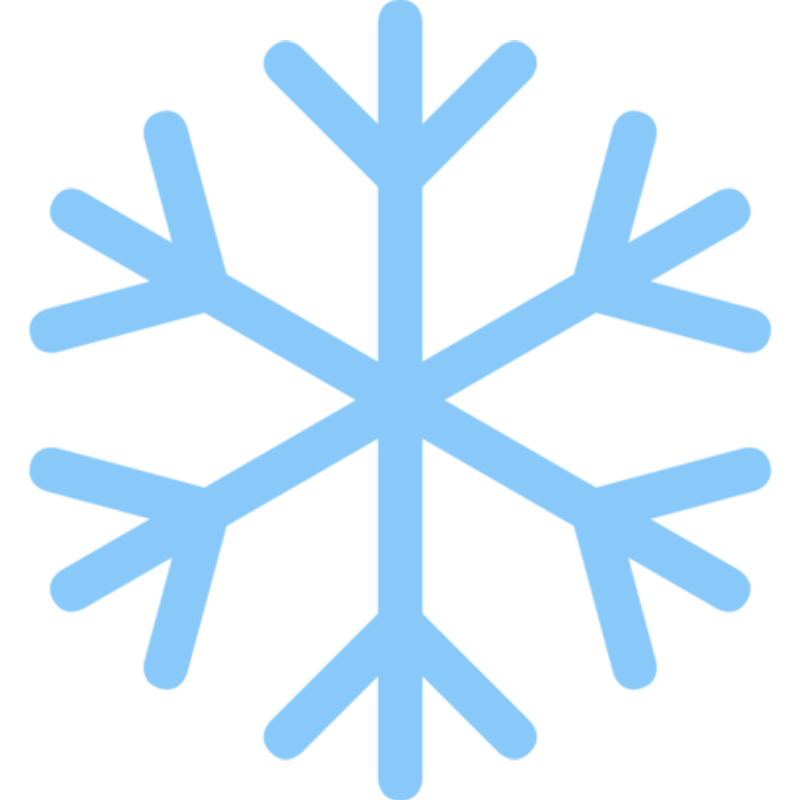}}).  }

    \label{fig:training}
\end{figure}

The proposed method~\methodname~employs a pre-trained CLIP model as $f_\theta$, consisting of an image encoder $\mathcal{E}_I$ and a text encoder $\mathcal{E}_T$, as shown in Fig.~\ref{fig:training}. This section details its main components: text-prompt conditioning, continual read-only prompts, and prediction ensembling.  

\subsection{Text-prompt conditioning}\label{sec:conditioning}
Contrary to previous DIL methods~\cite{wang2022sprompt,Nicolas_2024_WACV}, our objective is to design a tailored methodology for deepfake detection in an incremental setting.
To this extent, in detecting synthetic data the model should focus more on specific attributes of the image (\ie{} visual artifacts or inconsistencies) than on the image content.
However, it has been observed that CLIP focuses on spurious and core features when classifying an image~\cite{an2023more}.
Thus, ignoring such spurious correlations can force CLIP to pinpoint salient artifacts in the image, making it more robust in detecting deepfakes.
Similar to~\cite{an2023more}, we aim to focus CLIP attention on features that are more relevant for the synthetic content detection. We propose to infer the object's class in a zero-shot fashion using CLIP and to use such information to condition the textual prompts, during training and inference.
Nevertheless, the semantic space of image classes is usually unknown in deepfake detection datasets, thus we pre-define a set of classes $\mathcal{C}$. 
Given an input image $x$, we predict a category $c^\ast\in\mathcal{C}$ as: 
\begin{equation}
c^\ast=\argmax_{c\in\mathcal{C}}\text{sim}\left(\mathcal{E}_I\left({x}\right), \mathcal{E}_T\left({c}\right)\right)
\end{equation}

\noindent where $\text{sim}(\cdot, \cdot)$ is the cosine similarity,  computed as  $\text{sim}(u,v)=(u\cdot v)/\|u\| \|v\|$. In this step, we augment the class names with the textual prompt \texttt{``a photo of a \{\textcolor{TextBlue}{CLS}\}''}~\cite{radford2021learning}. 
Then, we use the predicted category to condition the textual prompt, obtaining as a result the prompt \texttt{``a \{\textcolor{TextGreen}{real}/\textcolor{TextRed}{fake}\} photo of a \{$c^\ast$\}''}. In practice, instead of using just the highest-scoring one, we consider the first top-$c$ predicted classes and get $c$ conditioned prompts. This is because the label set $\mathcal{C}$ is agnostic to the categories in the dataset sequence $\mathcal{D}$. By considering the top-$c$ predictions, we account for potential uncertainties and provide more contextual information about the input image. For ease of reading, we will consider $c=1$ in the following. The obtained conditioned textual prompts are then used for both training and inference time, as shown in Fig.~\ref{fig:enter-label}.

\subsection{Continual read-only prompts}
\label{sub:prompt-tuning}
To avoid solely relying on a single prompt while maintaining a low computational overhead, we propose to employ read-only prompts~\cite{lee2023read} as a substitute for prompt-tuning.
In particular, read-only prompts do not alter the internal representation of CLIP, and thus, at inference time we can concatenate prompts from different tasks and prepend them to the input.
Let $p_v^k\in\mathbb{R}^{L\times D_v}$ be the visual read-only prompt of task $k$, with length $L$ and embedding dimension $D_v$, and let $p_t^k\in\mathbb{R}^{L\times D_t}$ be the correspondent textual prompt for task $k$.
Then, at task $k$, these are prepended to the visual and text encoder input as described in Sec.~\ref{sub:fine-tuning}.
The output of both encoders is dropped except for prepended prompts, which are the only trainable parameters in our setting. 
To train prompts, we employ a contrastive cross-entropy loss. Specifically, if we consider the case of a fake image, the loss is computed for each pair of output prompts as follows:
\begin{equation}
    \mathcal{L}_{cce} = \frac{\exp(\overline{\text{sim}}(v^k, f^k))}{\exp(\overline{\text{sim}}(v^k, r^k)) + \exp(\overline{\text{sim}}(v^k, f^k))}
\end{equation}
where $v^k$, $r^k$, and $f^k\in\mathbb{R}^{L\times D}$ are the visual, real, and fake output prompts, and $\overline{\text{sim}}(\cdot, \cdot)$ is the average cosine similarity between text and visual prompts:
\begin{equation}
\label{eq:sim}
    \overline{\text{sim}}(v, f) = \frac{1}{L}\sum_{i=1}^L \frac{v_i \cdot f_i}{\|v_i\| \|f_i\|}
\end{equation}
Analogously, for a real image, the similarity at the numerator is computed between $v^k$ and $r^k$, while the denominator is unaltered.
As we mention in Sec.~\ref{sec:conditioning}, instead of considering just the top class predicted by CLIP, we use the top-$c$ ones.
As a result, we also average \eqref{eq:sim} across $c$ classes aside from the length dimension.

\subsection{Predictions ensembling}\label{sec:ensembling}
\begin{figure}
    \centering
    \includegraphics[width=\linewidth]{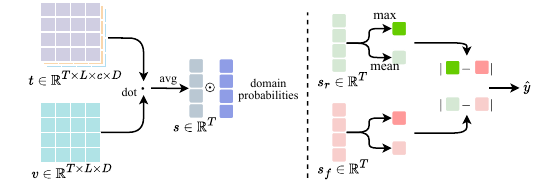}
    \caption{\textbf{Illustration of the ensembling.} We compute the average similarity from the visual and textual prompts $v$ and $t$ obtained from the respective encoders. Then we weight the scores with the domain probabilities. This is repeated for real and fake textual prompts. The obtained $s_r$ and $s_f$ are used to obtain the predicted class $\hat{y}$.}
    \label{fig:ensembling}
\end{figure}
As introduced in Sec.~\ref{sub:prompt-tuning}, at inference time our method does not require estimating the optimal set of parameters for each image compared to previous approaches.
Instead, \methodname can leverage properties of read-only prompts to avoid altering the internal representation of CLIP.
In practice, read-only prompts are unaware of other prompts in the forward pass, thus, prompts of different tasks do not influence their behavior.
Therefore, this allows us to concatenate prompts of all seen domains, where each will focus on aspects of the image that are salient for a specific image generator.
Formally, the input and output of the visual and text encoder are defined as follows:
\begin{align}
    \label{eq:v_p}
    &\mathcal{E}_I\left(\left[ \{p_v^k\}_{k=1}^T, x_{cls}, x_{img} \right]\right) = \left\{v^k\right\}_{k=1}^T \\
    \label{eq:v_r}
    &\mathcal{E}_T\left(\left[\{p_t^k\}_{k=1}^T, \text{a \texttt{\textcolor{TextGreen}{real}} photo of a } {\texttt{\textcolor{TextBlue}{CLS}}} \right]\right) = \left\{r^k\right\}_{k=1}^T \\
    \label{eq:v_f}
    &\mathcal{E}_T\left(\left[\{p_t^k\}_{k=1}^T, \text{a \texttt{\textcolor{TextRed}{fake}} photo of a } {\texttt{\textcolor{TextBlue}{CLS}}} \right]\right) = \left\{f^k\right\}_{k=1}^T
\end{align}
where \eqref{eq:v_p} shows the input and output of the visual encoder, and \eqref{eq:v_r} and \eqref{eq:v_f} respectively the real and fake input and output prompts of the text encoder.
To assign a score to a target image, we first use \eqref{eq:sim} to compute the average similarity for each pair of task prompts, $(v^k, r^k)$ and $(v^k, f^k)$.
As a result, we obtain two score vectors, $s_r\in\mathbb{R}^{T}$ and $s_f\in\mathbb{R}^{T}$, that respectively represent the prompts confidence in predicting whether the image is real or generated.
We scale each score vector entry by the likelihood that the generator corresponding to the entry has generated the image.
In practice, we follow previous works~\cite{wang2022sprompt,Nicolas_2024_WACV} and use a k-means classifier on the $\texttt{CLS}$ token of the image to extract the probability distribution, and scale scores vector entries by the computed likelihoods.
This allows for modulating the confidence of task prompts based on the likelihood that the image belongs to a specific domain.
The task scores with the highest magnitude usually lead to better predictions, however, when confidence is low, exploiting the decisions of all task parameters leads to better accuracies (refer to Tab.~\ref{tab:ablation3}).
Thus, we compute both the maximum and mean of predictions:
\begin{align}
    s_r^* = \max\{s^1_r, ..., s^T_r\},&\quad s_f^* = \max\{s^1_f, ..., s^T_f\}, \\
    \bar{s}_r = \frac{1}{T}\sum_{k=1}^T s_r^k,&\quad \bar{s}_f = \frac{1}{T}\sum_{k=1}^T s_f^k,
\end{align}
Then, for the final prediction of the model, we use the score with the maximum confidence if the relative maximum confidence, $\mid s_r^* - s_f^*\mid$ is greater than the relative mean confidence, $\mid \bar{s}_r - \bar{s}_r \mid$, otherwise, we uses the mean of logits:
\begin{equation}
\label{eq:maxmean}
    \hat{y} = \begin{cases}
        \argmax\{s_r^*, s_f^*\}, &\text{if} \mid s_r^* - s_f^*\mid \ge \mid \bar{s}_r - \bar{s}_f \mid \\
        \argmax\{\bar{s}_r, \bar{s}_f\}, &\text{otherwise}
    \end{cases}
\end{equation}
By analyzing the confidence of the predictions, \methodname can automatically decide whether to use a mixture of experts or the score with the highest confidence, improving performance.

\section{Experiments}\label{sec:exp}

\noindent\textbf{Dataset.} We perform experiments on the continual deepfake detection benchmark CDDB~\cite{li2023continual}. It gathers deepfakes from different generative models, gradually introduced to simulate the real-world deepfake's evolution. In particular, it designs three different evaluation setups, \ie, Easy, Hard, and Long. We select the most challenging, \ie, the Hard sequence task (CDDB-Hard) in order to be comparable with previous methods. In particular, it consists of learning on $5$ sequential deepfake detection domains, which are GauGAN~\cite{park2019semantic}, BigGAN~\cite{brock2018large}, WildDeepfake~\cite{zi2020wilddeepfake}, WhichFaceReal~\cite{whichfacewebsite}, and SAN~\cite{dai2019second} respectively.

\noindent\textbf{Metrics.} We perform the evaluation in terms of task-wise average accuracy (AA), which computes the average of all the task-based accuracies, as well as the average forgetting degree (AF), measuring the average decrease in accuracy on previous tasks after learning new tasks. 
In addition, we show the task-agnostic average accuracy (TAA), \ie{} the accuracy of predictions calculated over all the images without considering the task, at the end of the training.

\noindent\textbf{Implementation Details.} We use CLIP (ViT-B/16), therefore $D=512$. We set the length of both visual and textual read-only prompts as $L=7$ and use the top-$c$ classes with $c=5$. 
For the closed-set of categories $\mathcal{C}$ we use ImageNet-1k~\cite{russakovsky2015imagenet} classes for datasets containing images from general context and six handcrafted ones for face datasets. The six face classes are obtained as a cross product between $\{$young, middle-aged, old$\}$ and $\{$male, female$\}$.
We use the SGD optimizer with a learning rate of $0.01$ and cosine annealing with a constant warm-up of one epoch. We use $20$ epochs per domain. Input images are resized to a resolution of $224\times224$, and the data augmentation consists of horizontal flipping, random cropping, and color jittering.

\subsection{Comparative Results}

\begin{table}[ht]
    \begin{center}
    \begin{tabular}{l c c c c}
        \toprule
        \textbf{Method} & ~\textbf{Prompts}~ & \textbf{Buffer size} & ~\textbf{AA} $\uparrow$~ & \textbf{AF} $\uparrow$ \\
        \cmidrule{1-5}
            LRCIL~\cite{pellegrini2020latent} & $\times $ & \multirow{3}{*}{100 samples/class} & 76.39 & -4.39 \\
            iCaRL~\cite{rebuffi2017icarl} & $\times$ &             & 79.76 & -8.73 \\
            LUCIR~\cite{hou2019learning} & $\times$ &             & 82.53 & -5.34 \\ 
        \cmidrule{1-5}
            LRCIL~\cite{pellegrini2020latent}  & $\times$ & \multirow{4}{*}{50 samples/class} & 74.01 & -8.62 \\
            iCaRL~\cite{rebuffi2017icarl} & $\times$ &             & 73.98 & -14.50 \\
            LUCIR~\cite{hou2019learning}  & $\times$ &             & 80.77 & -7.85 \\
            DyTox~\cite{douillard2022dytox} & \checkmark &           & 86.21 & -1.55 \\ 
        \cmidrule{1-5}
            EWC~\cite{kirkpatrick2017overcoming} & $\times$ & \multirow{8}{*}{0 samples/class} & 50.59 & -42.62 \\
            LwF~\cite{li2017learning}   & $\times$ &             & 60.94 & -13.53 \\
            DyTox~\cite{douillard2022dytox} & \checkmark &  & 51.27 & -45.85 \\
            L2P~\cite{wang2022learning} & \checkmark &  & 61.28  & -9.23 \\
            S-iPrompts~\cite{wang2022sprompt} & \checkmark &  & 74.51 & -1.30 \\ 
            MoP-CLIP~\cite{Nicolas_2024_WACV} & \checkmark &  & 88.54 & -0.79 \\
            S-liPrompts~\cite{wang2022sprompt} & \checkmark &  & 88.65 & \textbf{-0.69} \\ 
            \rowcolor{TableBlue} \methodname & $\checkmark$ &     & \textbf{90.28} & -0.71 \\
        \bottomrule
    \end{tabular}
    \end{center}
    \caption{\textbf{Results on CDDB-Hard.} Blue is \inlineColorbox{TableBlue}{our method} and the best results are in \textbf{bold}. We also report if methods are prompt-based and their buffer size. }
    \label{tab:cddbhard}
\end{table}

\begin{figure}[!ht]
    \centering
    \includegraphics[width=.95\linewidth]{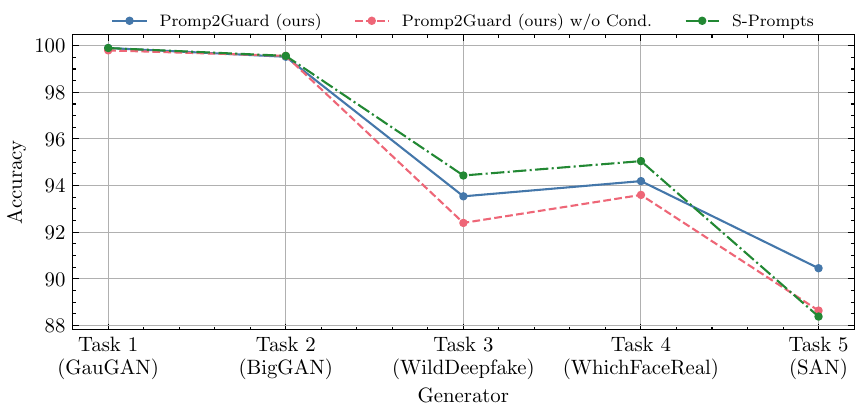}
    \caption{\textbf{Accuracy across tasks.} We show the task-wise average accuracy (AA) values for \inlineColorbox{TableBlue}{\methodname} and for the competitor \inlineColorbox{TableGreen}{S-Prompts} across all the tasks of CDDB-Hard. We also show \inlineColorbox{TablePink}{\methodname w/o conditioning}, \ie~without the step described in Sec.~\ref{sec:conditioning}. We plot the AA computed up to the $i$-th domain, against the domain index. 
    }
    \label{fig:acc_plot}
\end{figure}

We compare~\methodname against several state-of-the-art methods including:  non-prompting approaches such as EWC~\cite{kirkpatrick2017overcoming}, LwF~\cite{li2017learning}, LUCIR~\cite{hou2019learning}, iCaRL ~\cite{rebuffi2017icarl}, and LRCIL~\cite{pellegrini2020latent} and prompting-based methods such as L2P~\cite{wang2022learning}, DyTox~\cite{douillard2022dytox}, S-Prompts~\cite{wang2022sprompt}, MoP-CLIP~\cite{Nicolas_2024_WACV}. Our method is based on an exemplar-free DIL approach, thus we can assume as principal competitors EWC, LwF, DyTox, L2P, S-Prompts, and MoP-CLIP. 

Tab.~\ref{tab:cddbhard} presents the results on CDDB-Hard. 
The proposed~\methodname~outperforms previous methods, either exemplar-free or replay-based, delivering significantly better results in terms of AA. In particular, it surpasses the state-of-the-art S-Prompts by +1.63\% on the AA and achieves a low AF of -0.71\%.
Fig.~\ref{fig:acc_plot} further illustrates the AA curve on CDDB-Hard of the proposed method, with and without the text-prompt conditioning, and of the competitor S-Prompts. 

\begin{figure}
    \centering
    \vspace{-40pt}
    \includegraphics[width=.4\linewidth]{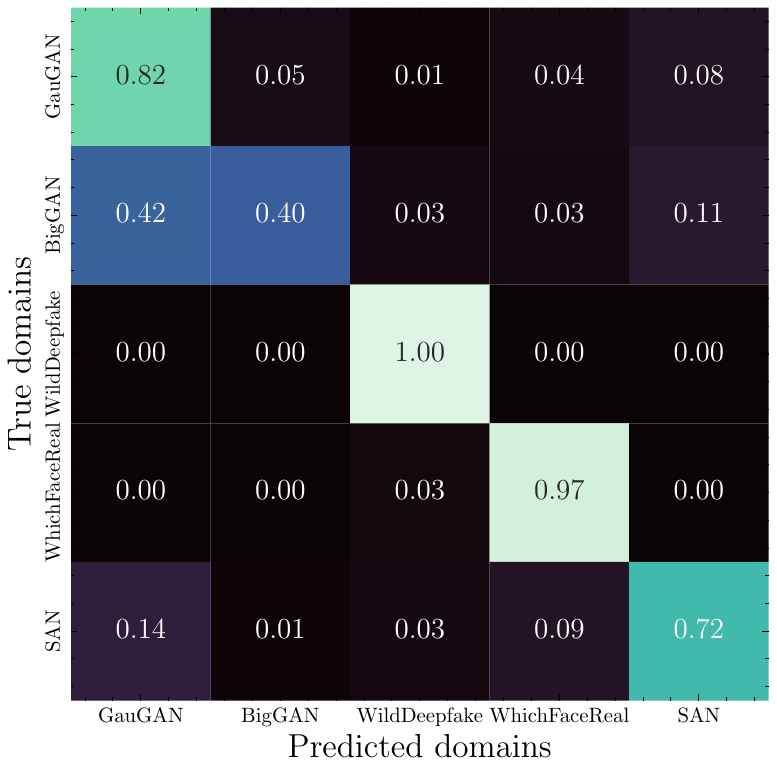}
    \caption{\textbf{Task Confusion.} Confusion matrix of the domain classification of the proposed \methodname on CDDB-Hard.}
    \label{fig:confmax}
\end{figure}

Tab.~\ref{tab:domain_details} shows the task-wise accuracy on each domain, the task-wise average accuracy (AA), and the task-agnostic average accuracy (TAA) of the proposed~\methodname~against the main competitor S-Prompts. The key insight is that our method achieves good accuracy across all the tasks, including the last and more challenging one, even if the accuracies in the previous tasks are slightly lower when compared to those of S-Prompts. In particular, S-Prompts obtains a task-wise accuracy of 68.89\% on the last domain, while our~\methodname~ gains a +12.22\% improvement. \methodname~benefits from the ensembling described in Sec.~\ref{sec:ensembling}, particularly on samples from the last domain SAN, which corresponds to low domain classification accuracy, as shown in the confusion matrix in Fig.~\ref{fig:confmax}. Despite BigGAN having the lowest domain classification accuracy, its task-wise accuracy remains high. Therefore the model leverages the ensembling and correctly classifies images as real or fake, even when the domain is misclassified.

\begin{table}[ht]
    \begin{center}
    \scriptsize
    \begin{tabular}{lccccccc}
    \toprule
        \multirow{2.5}{*}{\textbf{Method}} & \multicolumn{5}{c}{\textbf{Dataset}}&\multicolumn{2}{c}{\textbf{Metrics}}\\
    \cmidrule(lr){2-6}
    \cmidrule(lr){7-8}
        & ~GauGAN~ & BigGAN & ~WildDeepfake~ & WhichFaceReal & SAN & ~AA $\uparrow$~ & TAA $\uparrow$ \\
    \midrule
         S-Prompts \cite{wang2022sprompt} & \textbf{99.30}   & \textbf{96.75}  & \textbf{82.06} & \textbf{96.25} & 68.89 & 88.65 & \textbf{91.54} \\ 
         \rowcolor{TableBlue} \methodname & 98.70   & 94.38  & 81.73 & 95.50 & ~\textbf{81.11}~ & \textbf{90.28} & 90.98\\
    \bottomrule
    \end{tabular}
    \end{center}
    \caption{\textbf{Comparison of task-wise accuracy across different domains, AA and TAA (\%).} We show task-wise accuracy for each task of CDDB-Hard, both for S-Prompts and \inlineColorbox{TableBlue}{our proposed method}.}
    \label{tab:domain_details}
   
\end{table}

\begin{figure}
    \centering
    \includegraphics[width=\textwidth]{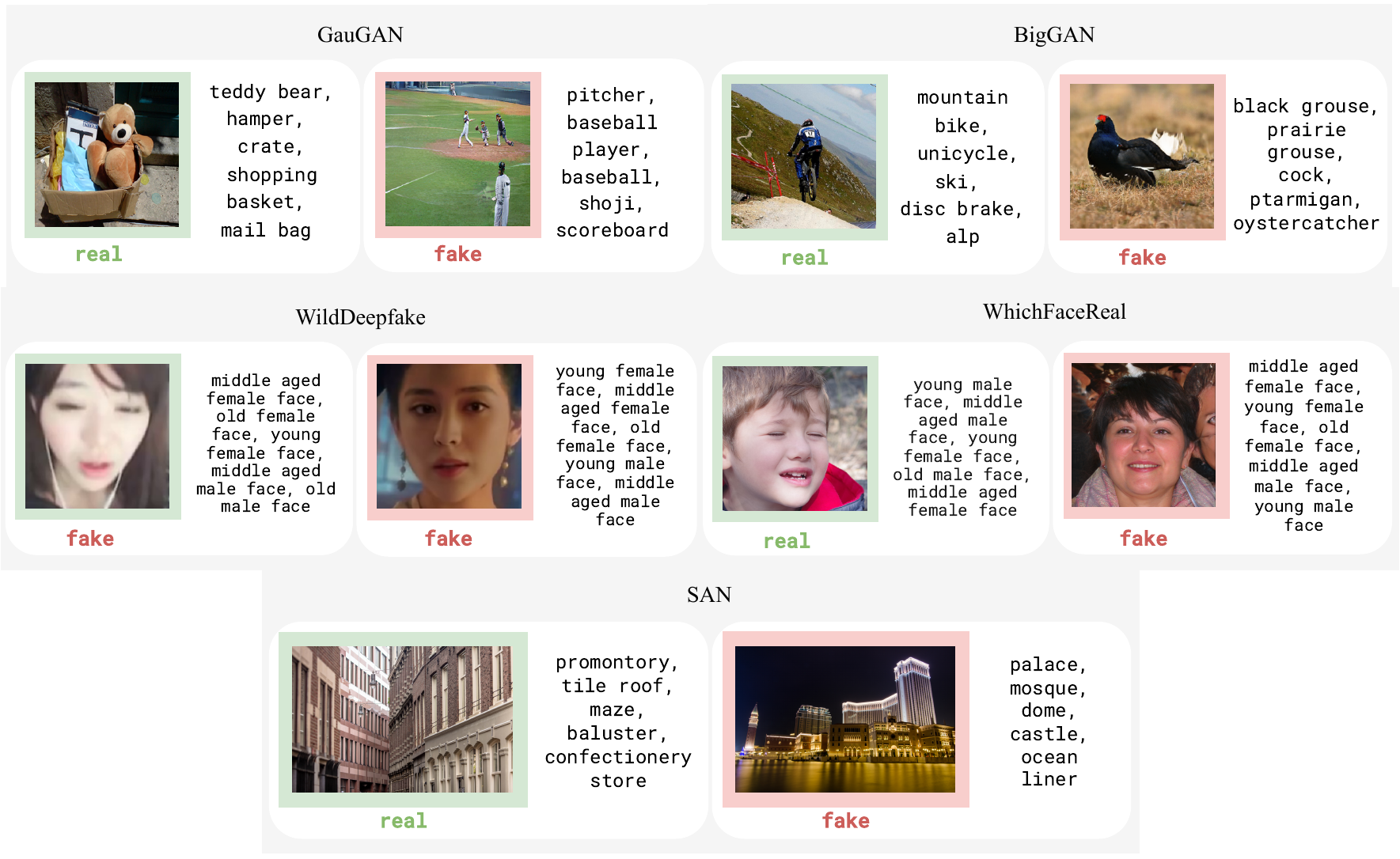}
    \caption{\textbf{Qualitative results.} We show the prediction of~\methodname~on test images from each task of CDDB. The colored frame around the image indicates the ground truth class, while the text underneath is the predicted one. We report on the right the top-$c$ classes predicted and used for text-prompt conditioning.}
    \label{fig:qualitatives}
\end{figure}

Fig.~\ref{fig:qualitatives} presents the qualitative results of our proposed~\methodname~in detecting deepfakes from all the tasks in CDDB-Hard. We report the top-$c$ predicted classes used for text-prompt conditioning, as described in Sec.~\ref{sec:conditioning}. The model is capable of detecting deepfakes for most of the cases. Moreover, the predicted classes are consistent with the content of the images. We observe that using more than one class is beneficial, specifically when more objects are present.

\subsection{Ablations}

We ablate the proposed method~\methodname~on CDDB-Hard to validate the effectiveness of our design choices. First, we analyze the text-prompt conditioning described in Sec.~\ref{sec:conditioning}, then the ensembling of the predictions detailed in  Sec.~\ref{sec:ensembling}. 

\noindent\textbf{Text-prompt Conditioning.} In Tab.~\ref{tab:ablation1} we assess the efficacy of conditioning the textual prompts on the category classified via zero-shot CLIP. When this step is added, we gain a $+1.64\%$ improvement in the AA and $+0.14\%$ in the AF. Our experiments validate the effectiveness of this choice, which lets the model focus more on salient artifacts relevant to deepfake detection rather than on the objects present in the image.
\begin{table}[ht]
    \begin{center}
        \begin{tabular}{l c c c}
            \toprule \textbf{Method} & ~\textbf{Text-prompt conditioning}~ & ~\textbf{AA $\uparrow$}~ & \textbf{AF $\uparrow$} \\
            \midrule
            \methodname & $\times$ &   88.64  & -0.85 \\
             \rowcolor{TableBlue}  \methodname & \checkmark  & \textbf{90.28}  & \textbf{-0.71} \\
            \bottomrule
        \end{tabular}
    \end{center}
    \caption{\textbf{Ablation on text-prompt conditioning.}  We report the results with and without conditioning the textual prompts on the category classified by zero-shot CLIP. Blue is \inlineColorbox{TableBlue}{our configuration}. }
    \label{tab:ablation1}
\end{table}

\noindent\textbf{Prediction ensembling.} In Tab.~\ref{tab:ablation3} we compare three different ensembling techniques that can be used to obtain the final prediction $\hat{y}$. Directly averaging the scores across the tasks (termed here as \textit{mean}) produces the worst results. 
Using always the scores with the maximum
confidence (\textit{max}) results in better performance. Lastly, the highest results, both in AA and AF, are achieved with the ensembling defined in~\eqref{eq:maxmean} (termed here as \textit{max} \& \textit{mean}). 
\begin{table}[ht]
    \begin{center}
        \begin{tabular}{l c c c}
            \toprule \textbf{Method} & ~\textbf{Prediction ensembling}~ & ~\textbf{AA $\uparrow$}~ & \textbf{AF $\uparrow$} \\
            \midrule
            \methodname & mean & 83.41 & -1.47 \\
            \methodname & max & 89.98 &  -1.15 \\ 
             \rowcolor{TableBlue}  \methodname & max \& mean & \textbf{90.28} & \textbf{-0.71} \\
            \bottomrule
        \end{tabular}
    \end{center}
    \caption{\textbf{Ablation on prediction ensembling.} We compare different ensembling choices for the prediction. Blue is \inlineColorbox{TableBlue}{our configuration}.}
    \label{tab:ablation3}
\end{table}

\section{Conclusions}

In this work we address the challenging problem of continual deepfake detection. We propose~\methodname, a novel exemplar-free solution that leverages VLMs, read-only multi-modal prompts, and a text-prompt conditioning specifically tailored to the task. 
Our experimental evaluation confirms the effectiveness of~\methodname~in achieving state-of-the-art results in task-wise average accuracy on the challenging CDDB benchmark. 
As future work, we plan to extend our method beyond the use of a closed label set, harnessing the power of vocabulary-free classification~\cite{conti2023vocabularyfree}, and to evaluate it on images coming from more recent generators, \eg~including diffusion-based deepfakes. 
Additionally, we plan to address the scalability limitations of our method, which are constrained by the token length limitation of the VLM’s text encoder. This restriction limits the number of domains that learnable prompts can represent at inference time, thereby constraining the number of tasks that can be handled simultaneously in a DIL setting.

\section*{Acknowledgments}

We acknowledge the CINECA award under the ISCRA initiative, for the availability of HPC
resources. This work was also sponsored by  PNRR FAIR - Future AI Research (PE00000013), funded by NextGeneration EU and supported by the EU project AI4TRUST (No.101070190). Thomas De Min is funded by NextGeneration EU.

{
    \small
    \bibliographystyle{splncs04}
    \bibliography{main}
}

\end{document}